\def\year{2020}\relax
\documentclass[letterpaper]{article} 
\usepackage{aaai20}  
\usepackage{times}  
\usepackage{helvet} 
\usepackage{courier}  
\usepackage[hyphens]{url}  
\usepackage{graphicx} 
\usepackage{graphicx}  
\title{Multi-tier Automated Planning for Adaptive Behavior (Extended Version)%
\thanks{Authors are listed in alphabetical order. 
}
}
\author{
Daniel~Ciolek \and Nicol\'as~D'Ippolito \\
	Departamento de Computaci\'on \\
	Universidad Nacional de Buenos Aires \\
	Argentina 
\And
Alberto Pozanco \\
	Departamento de Inform\'atica \\
	Universidad Carlos III de Madrid \\
	Spain
\And
Sebastian Sardi\~na \\
	School of Science \\
	RMIT University \\
	Australia
}

\usepackage{paralist}
\usepackage{enumitem}
\usepackage{helvet}
\usepackage{courier}

\usepackage{latexsym}
\usepackage{amsmath}
\usepackage{amssymb}
\usepackage{circle}
\usepackage{stmaryrd}

\usepackage{comment}
\usepackage{xspace}
\usepackage{soul}\setuldepth{x}
\usepackage{url}
\usepackage{filecontents}

\usepackage{newfloat}
\DeclareFloatingEnvironment[
   fileext=lol,
   name=Listing
]{listing}

\usepackage{subcaption}
\DeclareCaptionSubType{listing}

\usepackage[table,xcdraw,usenames,dvipsnames]{xcolor}

\usepackage[nameinlink]{cleveref}	

\newcommand{\mathname}[1]{\ensuremath{\text{\textit{#1}}}}
\newcommand{\propername}[1]{\text{\textsf{\small \smash{#1}}}\xspace}
\newcommand{\name}[1]{\mathname{#1}}

\newcommand{\actionfont}[1]{\text{\textsc{#1}}}

\newcommand{\defterm}[1]{\ul{\emph{#1}}}
\renewcommand{\defterm}[1]{\emph{#1}}
\renewcommand{\defterm}[1]{\emph{\textbf{#1}}}

\newcommand{\C}{\mathcal{C}} \newcommand{\D}{\mathcal{D}}
 
\newcommand{\G}{\mathcal{G}}

\newcommand{\M}{\mathcal{M}} 
 \renewcommand{\P}{\mathcal{P}}

\newcommand{\tup}[1]{\langle #1\rangle}            
\newcommand{\tuple}[1]{\tup{#1}}            

\newcommand{\set}[1]{\{#1\}}                      
\newcommand{\compl}[1]{\overline{#1}}

\newcommand{\Goal}{\phi_{\text{goal}}}

\newcommand{\last}{\name{last}}

\newcommand{\Pre}{\name{Pre}}
\newcommand{\Eff}{\name{Eff}}

\newcommand{\FOND}{FOND\xspace}

\newcommand{\PDDL}{\propername{PDDL}}

\newcommand{\citeby}[1]{\shortciteA{#1}}
\renewcommand{\citeby}[1]{\citeauthor{#1}~\shortcite{#1}}

\newcommand{\strongcyclic}{strong-cyclic\xspace}

\newcommand{\ceffect}[2]{#1 \Rightarrow #2}
\newcommand{\DT}{\D^\top}
\renewcommand{\DT}{\hat{\D}}

\usepackage[amsmath,thmmarks]{ntheorem}
\theoremseparator{.}

\theorembodyfont{\itshape}
\newtheorem{theorem}{Theorem}

\newtheorem{lemma}{Lemma}

\theorembodyfont{\normalfont}
\theoremsymbol{\ensuremath{\blacksquare}}
\newtheorem{definition}{Definition}

\theoremsymbol{\ensuremath{\square}}
\newtheorem{example}{Example}

\theorembodyfont{\normalfont}
\theoremstyle{nonumberplain}
\theoremsymbol{\ensuremath{\square}}
\theoremheaderfont{\normalfont\itshape}
\newtheorem{proofsk}{Proof Sketch}

\usepackage[textsize=scriptsize,textwidth=1.7cm]{todonotes}
\usepackage{cooltooltips}
\usepackage{etoolbox}
\makeatletter
\patchcmd{\@addmarginpar}{\ifodd\c@page}{\ifodd\c@page\@tempcnta\m@ne}{}{}
\makeatother
\reversemarginpar

\usepackage{listings}

\lstdefinelanguage{pddl}
{
    keywords={
        define,
    },
    keywordstyle=[2]{\color{red}},
    morekeywords={[2]
        domain,
        problem,
        requirements,
        predicates,
        types,
        objects,
        action,
        init,
        goal,
    },
    keywordstyle=[3]{\color{MidnightBlue}},
    morekeywords={[3]
        parameters,
        vars,
        precondition,
        effect,
    },
    keywordstyle={[4]\color{OliveGreen}},
    morekeywords={[4]
        forall,
        or,
        and,
        not,
        when,
        =,
        exists,
        imply,
        oneof
    },
    comment=[l]{\;},
    sensitive=true
}[keywords, comments]

\lstdefinelanguage[mapl]{pddl}[]{pddl}
{
    morekeywords={[2]
      sensor,
    },
    morekeywords={[3]
      sense,
      replan
    },
    morekeywords={[4]
      KIF,
    },
}

\newcommand{\pddl}[1]{\text{\lstinline[basicstyle=\footnotesize\ttfamily]{#1}}}

\usepackage{etoolbox}
\makeatletter
\patchcmd\@combinedblfloats{\box\@outputbox}{\unvbox\@outputbox}{}{\errmessage{\noexpand patch failed}}
\makeatother

\usepackage{tikz}
\usepackage{pgfplots}

\usetikzlibrary{petri,arrows,decorations,backgrounds,matrix,automata,shapes,shadows,patterns,fit,calc}

\tikzstyle{every picture}=[->,>=stealth',shorten >=1pt,auto,node distance=1.3cm,semithick]  

\tikzstyle{place}=[circle,thick,draw=black,minimum size=4mm]
\tikzstyle{invisible place}=[place,draw=none,fill=none]
\tikzstyle{transition}=[rectangle,inner ysep=1,thick,draw=black!75,fill=black!10,minimum size=2mm,,minimum width=4mm]
\tikzstyle{itransition}=[transition,draw=none,fill=none]
  \tikzstyle{every label}=[black]
\tikzstyle{every initial by arrow}=[initial text=]
\tikzstyle{every state}=[fill=none,draw=black,text=black,inner sep=1pt,minimum size=7mm]
  
\usetikzlibrary{external}
\tikzsetexternalprefix{externalfigs/}

\makeatletter
\AtBeginDocument{\@ifpackageloaded{beamer}{%
	\newcommand*{\overlaynumber}{\number\beamer@slideinframe}
	\tikzset{
	  beamer externalizing/.style={%
	    execute at end picture={%
	      \tikzifexternalizing{%
	        \ifbeamer@anotherslide
	        \pgfexternalstorecommand{\string\global\string\beamer@anotherslidetrue}%
	        \fi
	      }{}%
	    }%
	  },
	  external/optimize=false
	}
	\let\orig@tikzsetnextfilename=\tikzsetnextfilename
	\renewcommand\tikzsetnextfilename[1]{\orig@tikzsetnextfilename{#1-\overlaynumber}}
	\tikzset{every picture/.style={beamer externalizing}}
}
{\relax}
}
\makeatother

\begin{document}

\maketitle
\begin{abstract}
	A planning domain, as any model, is never “complete” and inevitably makes assumptions on the environment's dynamic. 
	By allowing the specification of just one domain model, the knowledge engineer is only able to make one set of assumptions, and to specify a single objective-goal.
	Borrowing from work in Software Engineering, we propose a \textit{multi-tier} framework for planning that allows the specification of different sets of assumptions, and of different corresponding objectives.
	The framework aims to support the synthesis of \emph{adaptive} behavior so as to mitigate the intrinsic risk in any planning modeling task.
	After defining the multi-tier planning task and its solution concept, we show how to solve problem instances by a succinct compilation to a form of non-deterministic planning.
	In doing so, our technique justifies the applicability of planning with \textit{both} fair and unfair actions, and the need for more efforts in developing planning systems supporting dual fairness assumptions. 
\end{abstract}

\section{Introduction}\label{sec:intro}

In AI planning~\cite{GhallabNauTraverso:BOOK04-Planning,GeffnerBonet:BOOK13-planning}, a plan is synthesized against \emph{a model} of the environment---a \emph{planning domain}---to achieve a given goal from an initial state of the environment.
Such model describes how actions change the world, via the specification of their preconditions and effects.
As any model, planning domains are never ``complete'' and they inevitable make assumptions on the dynamics of the environment.
A limitation of standard planning formalism is that they do not account for deviations from such assumptions, and hence are not well prepared for integration within an execution framework. 
A common approach to handle discrepancies between what the planner expected and what happened at run-time is to simply perform \emph{re-planning} or \emph{plan-repair}~\cite{Fox:ICAPS06}. 
But, why would the system keep reasoning about the \emph{same} model of the world that has been proven ``wrong''?

As an actor's view on planning becomes more prominent in the field~\cite{Ghallab.etal:AIJ14}, and inspired by work in Software Engineering~\cite{DIppolito.etal:ICSE14}, we propose a ``generalized''\footnote{Not in the sense of plans solving set of problem instances~\cite{Siddharth.etal:TR08-GENPLANNING}, but as generalization of the problem definition.} planning framework that aims to better account for the uncertainties at design/engineering time.
Concretely, rather than fixing the level of risk and objectives, we envision the specification of various assumption levels, as well as different goals for each assumption level. 
This is achieved by allowing the knowledge engineer to specify a family of planning domains, each carrying a set of assumptions.
For example, in an fully idealized model of the blocks world, a robotic arm always successfully grabs blocks, whereas in less idealized models the griper may fail when picking, maybe missing or even breaking it. 
Depending on the assumptions imposed on the gripper operation, one may aim for different types of block towers.

The aim of the framework is to synthesize not one but a \emph{collection} of inter-related policies embedding, together, \emph{adaptive} behavior. 
So, as the environment violates assumptions, the agent should gracefully ``degrade'' to less refined planning models.
Since such models carry less assumptions on the environment, less functionalities can generally be offered.
If the gripper may break a block while picking it, building a complete block may just not be achievable.
So, with model degradation, comes goal degradation, usually to a less ambitious (and often less demanding) one.

We call the above framework \emph{multi-tier adaptive planning} and is the topic this paper.
Let us start with a simple example to motivate the work and upcoming technical development.

\section{Running Example}\label{sec:example}
 \begin{listing*}
     \begin{sublisting}{.3\linewidth}
       \begin{lstlisting}[language=pddl]

(:action walk
  :parameters 
    (?o - Cell ?d - Cell)
  :precondition (and (at ?o) 
    (adj ?o ?d) (not (broken)))
  :effect (and 
    (not (at ?o)) (at ?d)) )

(:action run
  :precondition 
    (and (at c2) (not (broken)))
  :effect (and 
    (not (at c2)) (at c0)) )
		  
(:goal (and (at c0) 
            (not (scratch)) 
            (not (broken)) ))
       \end{lstlisting}
       \caption{In model $\D_3$, any running and walking always succeeds.}
     \end{sublisting}
     \begin{sublisting}{.3\linewidth}
       \begin{lstlisting}%[gobble=8]

(:action walk
 :parameters 
   (?o - Cell ?d - Cell)
 :precondition (and (at ?o)
   (adj ?o ?d) (not (broken)))
 :effect (oneof
   (and (not (at ?o)) (at ?d))
   (and (not (at ?o)) (at ?d) (scratch))) )

(:action run
  :precondition 
    (and (at c2) (not (broken)))
  :effect (oneof
    (and (not (at c2)) (at c0))
    (and (not (at c2)) (at c0) (scratch)) ))

(:goal (and (at c0) (not (broken))) )
       \end{lstlisting}
       \caption{In model $\D_2$, agent may move successfully but suffer minor scratch damage.}
     \end{sublisting}
     \qquad\quad
     \begin{sublisting}{.3\linewidth}
       \begin{lstlisting}%[gobble=8]

(:action walk
  :parameters (?o - Cell ?d - Cell)
  :precondition (and (at ?o)
    (adj ?o ?d) (not (broken)))
  :effect (oneof
    (and (not (at ?o)) (at ?d))
    (and (not (at ?o)) (at ?d) (scratch))
    (scratch) ))

(:action run
  :precondition 
    (and (at c2) (not (broken)))
  :effect (oneof
    (and (not (at c2)) (at c0))
    (and (not (at c2)) (at c0) (scratch))
    (broken) ))

(:goal (and (at c2) (not broken)) )
       \end{lstlisting}
       \caption{In model $\D_1$, movements may actually fail and may even leave the robot broken.}
     \end{sublisting}
     \vspace*{-0.2cm}
\caption{Actions walk left and run left in the three models.}
\label{lst:norunning_example_pddl}
\end{listing*}

%

\newcommand{\Pick}{\textsc{Pick}\xspace}
\newcommand{\Put}{\textsc{Put}\xspace}
\newcommand{\Glue}{\textsc{Glue}\xspace}

Consider a robot moving in a $1 \times n$ grid similar to the dust-cleaning robot example in~\cite{BonetGeffner:IJCAI15}.
The robot can \emph{walk} one cell at a time or it can \emph{run} covering multiple cells in one shot.
Unfortunately, the physical shape of the corridor is not fully known to the designer and the physical guarantees of the robot's actuators are not be fully known to the designer. 
Because of that, in some scenarios, some cells may be impossible to navigate and the robot may get damaged or even broken.
So, the knowledge engineer considers various possible assumption levels on the environment's behavior, together with corresponding adequate objectives.

In the most idealized model $\D_3$, the designer assumes that both walking and running actions succeed with no negative side-effects. The goal there is for the robot to reach a destination cell \pddl{c0} and intact. 
In a less idealized $\D_2$, both running and walking actions still cause the robot to advance, but no assumption can be made on their side effects and movement may cause minor damages. The robot should then just aim to reach the target destination \pddl{c0}.
Finally, in the least idealized $\D_1$, a walking action may sometimes cause the robot to get minor damages without even advancing, and even worse, a running action may get the robot \emph{broken} and render it unusable. Under those weaker assumptions, the robot should return to base location \pddl{c2} for servicing.

Under the above multi-tier specification, the robot tries its best, but adapts its behavior as it discovers some assumptions may not hold.
To do so, the robot initially assumes the most idealized world model and thus works for the most ambitious goal: reach destination undamaged.
But, upon observing an inconsistency with the assumptions, it must \emph{adapt} both the model of the environment as well as the objective being pursued.
For example, if the robot succeeds in advancing when moving but gets a minor damage, it should \emph{degrade} to $\D_2$. If it actually fails to move at all, it should \textit{degrade} to $\D_1$ to operate under such level weaker assumptions.

A solution to this scenario must, on the one hand, strive for the best possible solution and, on the other hand, be open to a potential future degradation.
Concretely, the robot should never attempt to perform an action that may prevent graceful degradation.
In our example, while, in principle, running would be the most efficient way to reach the destination, it may cause a catastrophic failure in tier level 1, precluding the goals of every tier.
Thus, the robot must be conservative and should cautiously move by walking.

In what follows, we propose a framework to specify problems like this one in the realms of automated planning, define its solution concept, and show how to compute solutions with existing planning technology.

\section{Multi-tier Planning}\label{sec:mtp_framework}
\newcommand{\HRFP}{\MTP}
\newcommand{\MTP}{\propername{MTP}}
\newcommand{\MTD}{\propername{MTD}}
\newcommand{\MTC}{\propername{MTC}}

\newcommand{\up}[1]{\ensuremath{{\uparrow}#1}}
\newcommand{\down}[1]{\ensuremath{{\downarrow}#1}}
\newcommand{\ups}[1]{\ensuremath{{\overline{\uparrow}}#1}}
\newcommand{\downs}[1]{\ensuremath{{\overline{\downarrow}}#1}}

\newcommand{\RUN}{\mathname{Runs}}
\newcommand{\FRUN}{\mathname{FairRuns}}
\newcommand{\InitStates}{\mathname{InitTrg}}
\renewcommand{\InitStates}{\mathname{Init}}
\newcommand{\Sup}{\mathname{Sup}}

In this section we propose a multi-tier automated planning framework in which the knowledge engineer can specify a ranked set of assumptions about the environment and a corresponding set of objective goals.
A solution to such framework will display \emph{adaptive} behavior at execution time, by aligning the model and objectives w.r.t. run-time observations.
Before doing so, though, we first go over the standard technical machinery for \emph{non-deterministic} planning.

A \defterm{fully observable non-deterministic (FOND) planning domain}~\cite{Rintanen:ECAI08,GerviniBonetGivan:IPC5} is a pair $\D = \tup{V, O}$ consisting of a set of Boolean state variables $V$ and an operator set $O$. 
A \defterm{state} $s \in 2^V$ is the set of propositional variables that are true in the state.
We use $S$ to denote the set of all states and $\compl{l}$ to denote the complement of literal $l$.

An \defterm{operator} is a tuple $\tup{o, \Pre_o,\Eff_o}$, where $o$ is a unique name, $\Pre_o$ is a Boolean condition over $V$ describing the \emph{preconditions} of operator $o$, and $\Eff_o = e_1 \mid \cdots \mid e_n$,  with $n \geq 1$, is the (non-deterministic) \emph{effect} of $o$ where each $e_i$ is a (set of) conditional effects $\ceffect{C}{E}$ with $C$ being a Boolean condition over $V$ and $E$ a set (conjunction) of literals.\footnote{This formalization of (non-deterministic) actions corresponds to 1ND Normal Form~\cite{Rintanen:ICAPS03} with no nested conditional (deterministic) effects, and to the usual \lstinline$(oneof e1 ... en)$ clauses in \PDDL~\cite{GerviniBonetGivan:IPC5} to model non-deterministic actions.} 
The intended meaning is that \emph{one} of the $e_i$ effects ensues non-deterministically, by the environment's choice. 

\newcommand{\Executions}{\mathname{Ex}}

A \defterm{policy controller} is a function $\pi : S \mapsto 2^O$ that maps state $s \in S$ to a set of (executable) actions $\pi(s)$.
A policy $\C$ executed from state $s \in S$ on domain $\D$ defines a set of \defterm{possible executions} $\Executions_\pi(\D,s)$ of the form $\lambda = s_0o_0s_1\cdots s_io_is_{i+1}\cdots$, where $s_0=s$, $o_i \in \pi(s_i)$, $s_i \models \Pre_{o_i}$, and $s_{i+1}$ is a possible successor state when $o_i$ is executed in state $s_i$ w.r.t. domain $\D$, for all $i \geq 0$.
We use $\last(\lambda)$ to denote the last state in (finite) execution $\lambda$ and $\Executions(\D,s)$ to the set of all possible executions in $\D$ from state $s \in S$ (i.e., $\Executions(\D,s) = \Executions_{\pi^*}(\D,s)$, where $\pi^*(s) = O$).

Finally, a \defterm{FOND planning problem} $\P = \tup{\D,s_I,\Goal}$ consists of a FOND  domain $\D$, an initial state $s_I$, and a goal $\Goal$ as a conjunction of literals from $V$. 
There has been several solution concepts for FOND planning depending on the fairness of non-deterministic actions.
Roughly speaking, a \defterm{fair action} is one in which all effects occur infinitively often when the action is executed infinitively many times in the same state~\cite{GeffnerBonet:BOOK13-planning,SardinaDippolito:IJCAI15}.
When all actions are assumed fair, a \defterm{\strongcyclic} plan guarantees that the agent, by ``re-trying,'' eventually achieves the goal~\cite{Cimatti.etal:AIJ03}.
In turn, when no fairness can be assumed, a plan with acyclic executions that reaches the goal in a bounded number of steps---a \defterm{strong} policy---is required. 
The \defterm{Dual FOND} (or FOND+) hybrid variation has recently been introduced to deal with domains that have both fair and unfair actions/effects~\cite{CamachoMcIlraith:KNOWPROS16,GeffnerGeffner:ICAPS18-FONDSAT}.
In that setting, a solution amounts to a policy whose ``fair'' executions w.r.t. the actions/effects assumed to be fair (not necessarily all) are goal reaching.
Lastly, we note that while planning under non-determinism is EXPTIME-complete~\cite{Rintanen:ICAPS04}, effective optimized techniques and solvers have been developed, and is an area of significant active work (e.g.,~\cite{Muise.etal:ICAPS12-PRP,Kuter.etal:ICAPS08-NDP,KissmannEdelkamp:KI09-GAMER,Muise.etal:AAAI14,GeffnerGeffner:ICAPS18-FONDSAT}).

%
%
%
%

\medskip
With the technical machinery on FOND planning at hand, we are ready to formally present our framework for multi-tier adaptive planning.
Following \cite{DIppolito.etal:ICSE14}, we aim for the knowledge engineering to be able to specify a variety of models carrying different assumptions.

\begin{definition}
A \defterm{multi-tier planning domain} (\MTD) is a tuple $\tuple{\Omega, \leq}$ such that:
\begin{enumerate}
    \item $\Omega$ is a set of FOND planning domains over the same variables $V$ and operator signatures, and every operator has the same preconditions across all domains in $\Omega$;

	\item $\leq$ is a partial-order relation over $\Omega$ such that $\D_1 \leq \D_2$ implies $\Executions(\D_2,s) \subseteq \Executions(\D_1,s)$ for all states $s \in S$; and
	
	\item $\leq$ has a greatest element in $\Omega$, denoted $\DT$, as well as a minimum element. 
	
\end{enumerate}
\end{definition}

The first condition states that an \MTD is just a collection of planning domains over the same vocabulary.
While the action names and their preconditions ought to be the same in all domains, the effects of each operator may differ across domains.
Such differences in operator effects will reflect \textit{different assumptions on the environment}.
However, the differences cannot be arbitrary: they are to reflect model ``refinements.''
This is achieved by the second condition, which specifies that domains in lower tiers of the hierarchy ($\D_1$) must produce the same behaviors as higher models ($\D_2$), and possibly more.
The intuition is that higher-level models are ``refinements" of lower-level models, posing possibly  more assumptions on the environment (e.g., by actions having fewer non-deterministic effects), hence permitting fewer execution runs.


As in standard planning, a problem instance task adds a specific initial situation and a (set of) objectives.

\begin{definition}
A \defterm{multi-tier planning problem} (\MTP) is a tuple $\M = \tuple{\tuple{\Omega, \leq}, s_I, \G}$ where $\tuple{\Omega, \leq}$ is an \MTD, $s_I$ is $\M$'s initial state, and $\G$ is a function mapping each domain $\D$ in $\Omega$ to a goal $\G(\D)$ (or just $\G_\D$). 
\end{definition}

Observe that unlike standard planning approaches, we allow the designer to specify various goals, depending on the risk imposed by the assumptions on the environment.
Often, the weaker the assumptions, the lower level of functionality that may be guaranteed~\cite{DIppolito.etal:ICSE14}.

\begin{example}
Listing~\ref{lst:norunning_example_pddl} depicts an \MTP for the no-running example of Section~\ref{sec:example}, in which the modeler specifies three planning domains.
The two actions \texttt{walk} and \texttt{run} are modeled at each of the three tiers.
The actions' schema and preconditions are the same across all tiers.
However, the effects differ progressively, as \textit{assumptions are relaxed} from higher tiers to lower tiers. 
For example, in third tier model $\D_3$, the most idealized model, it is assumed that movement actions always succeed, whereas in the lowest tier model $\D_1$, actions may fail and may even do so catastrophically. 
Also observe the goals for each tier may differ, since when assumptions are relaxed, more ambitious goals may not be achievable.  
Because effects are incrementally relaxed across tiers, runs at lowers tiers are strict super sets of those in upper tiers.
\end{example}

Finally, we define a structure that associates a specific policy to each domain in an \MTD, prescribing what behavior should ensue from the executor under the different models.

\begin{definition}
A \defterm{multi-tier controller} (\MTC) for an \MTD $\tuple{\Omega, \leq}$ is a function $\C: \Omega \mapsto (S \mapsto 2^{O})$ mapping each domain $\D \in \Omega$ to a specific policy $\C(\D)$ (or just $\C_\D$).
\end{definition}

The challenge now is to formally capture when an \MTC amounts to a ``solution'' strategy for a multi-tier planning problem.
To do so, it is important to first understand how the \MTC structure is meant to be deployed in the environment.
Intuitively, at any point in time, the executor is operating relative on some planning domain (i.e., model) of the world $\D$ from the ones available in $\Omega$, by carrying out its corresponding policy $\C(\D)$ so as to bring about the level's goal $\G(\D)$. 
Initially, the executor deploys policy $\C(\DT)$ from the initial problem state $s_I$ on the most idealized domain $\DT$, aiming at achieving the most ambitious goal $\G(\DT)$.
However, if at any point during execution, an inconsistency with the current model $\D_i$ is observed, the executor ought to switch to an alternative domain $\D_j \in \Omega$ such that $\D_j \leq \D_i$.
Technically, an inconsistency amounts to observing an actual state $s$ that cannot be explained with planning domain $\D_i$. 
Of course, once the executor switches downwards---referred as \textit{degradation}---the model it operates on to a more permissive one (i.e., one with weaker assumptions), the objective sought, and hence the strategy, must be changed too.
A smart executor, though, aims to degrade gracefully, that is, as little as possible, switching to a planning domain that retains as many assumptions as possible about the environment (and the most ambitious goal).

\smallskip
Let us now develop the solution concept for {\MTP}s.
We define the set of triggering states for a domain in an \MTD as those states in which the executor, when deployed in a given multi-tier controller as per the above operational scheme, may need to start operating under such domain.
As expected, the initial state $s_I$ is the triggering state for the highest level, most idealized, domain $\DT$.
For other domains, a triggering state amounts to a degradation step.

\begin{definition}
Let $\C$ be an \MTC for a \MTD $\tuple{\Omega, \leq}$, and let $s$ be a state (over variables $V_\Omega$).
We inductively define the set of \defterm{triggering initial states} for each planning domain $\D \in \Omega$ under $\C$, denoted $\InitStates(\D,\C)$, as follows:
\begin{enumerate}
     \item $\InitStates(\DT, \C) = \set{s_I}$;
     
     \item if $\D$ is not the maximum in $\Omega$ (i.e., $\D \not= \DT$), then
\[
\begin{array}{l}
	\InitStates(\D, \C) = \\{}
\phantom{xxx}	
	\{  
	s \mid
		\D < \D', s' \in \InitStates(\D'),  \\{}
\phantom{xxxxxx}
		\lambda \in \Executions_{\C_{\D'}}(\D',s'), o = \C_{\D'}(\last(\lambda)), \\{}		
\phantom{xxxxxx} \displaystyle
		\lambda o s \in \Executions(\D,s') \setminus 
					\bigcup_{\D'' : \D < \D''} \Executions(\D'',s'). 
	\}
\end{array}
\]
\end{enumerate}
\end{definition}

Let us explain the second case.
Suppose the executor has so far been carrying out policy $\C_{\D'}$ on a domain model $\D'$, from some (triggering) state  $s'$.
Suppose this has yielded execution run $\lambda$ (consistent with $\D'$).
However, when executing the next prescribed operator $o$ (as per the corresponding policy $\C_{\D'}$ for $\D'$), the resulting evolution to a state $s$ yields an execution $\lambda o s$ that can be explained by (i.e., its a legal execution in) domain $\D$ but not by any model higher than $\D$ (including $\D'$).
When this happens, state $s$ is a triggering state for $\D$, that is state where the executor may have to start operating under domain model $\D$ when using policy $\C_\D$.

Next, for a controller $\C$ to be a solution for an \MTP $\M$, it must achieve the associated goal of a domain in $\M$ \textit{from all the triggering states of the domain in question}.

\begin{definition}\label{def:mtc_solution}
An \MTC $\C$ is a \defterm{solution controller} for an \MTP $\M = \tuple{\tuple{\Omega, \leq}, s_I, \G}$  \emph{iff} for every domain $\D \in \Omega$, the projected policy $\C_\D$ is a solution plan for planning problem $\tuple{\D,s,\G_\D}$, for every state $s \in \InitStates(\D,\C)$. 
%
%
\end{definition}

Note that, unlike standard planning, this definition requires each policy to work from more than one initial state. 
However, it is not the case that all policies in $\C$ need to work from the initial state $s_I$. Such a requirement would be too demanding for capturing the intended operational framework as described above, this is because most policies, if not all but $\C_{\DT}$, will ever be used at state $s_I$ (unless the system comes back to such state after some degradation).

\section{Solving Multi-tier Planning Problems}\label{sec:solving_mtp}

\newcommand{\step}{\mathname{act}}
\newcommand{\pend}{\mathname{end}}
\newcommand{\unfair}{\mathname{unfair}}

\newcommand{\ounf}{o_\unfair}

\newcommand{\degrade}{\actionfont{degrade}}
\newcommand{\degrades}{\actionfont{deg}}
\newcommand{\continue}{\actionfont{continue}}
\newcommand{\continues}{\actionfont{cont}}
\newcommand{\mustDeg}{\mathname{mustDegrade}}

\newcommand{\isAligned}{\mathname{aligned}}

\newcommand{\checkgoal}{\actionfont{checkgoal}}

\newcommand{\checkgoals}{\actionfont{check}}

Informally, an \MTP is a collection of similar planning problems and a solution amounts to solution policies for each problem that can be ``connected'', if necessary, at degradation time.
A naive approach thus would repetitively compute solution policies for each planning problem, making sure they ``connect."
We show here we can solve the whole problem in a principled manner and in one shot.
Concretely, we build a single Dual \FOND planning task $\P_\M$ from a given \MTP $\M$ such that a strong-cyclic solution for $\P_\M$ amount to an \MTC solution for $\M$.
To argue for technique's generality, we first identify a meaningful fragment of \MTD{s}.

\begin{definition}
A planning domain $\D_2 = \tuple{V_2,O_2}$ is an \defterm{oneof-refinement} of a domain $\D_1 = \tuple{V_1,O_1}$ iff $V_1=V_2$ and for every $\tuple{o,\Pre^2_o,\Eff^2_o} \in O_2$, there is a $\D_1$-operator $\tuple{o,\Pre^1_o,\Eff^1_o} {\in} O_1$ such that $\Pre^1_o {=} \Pre^2_o$ and $\Eff^2_o {\subseteq} \Eff^1_o$;
\end{definition}

That is, $\D_2$ is like $\D_1$ but may contain fewer non-deterministic effects on some operators.
It turns out that, in the context of Dual \FOND planning, oneof-refinements capture all possible refinements in a multi-tier planning task---any \MTD is equivalent to a oneof-refinement type.

\begin{theorem}\label{theo:nd-refinment_universal}
Let $\tuple{\Omega,\leq}$ be an \MTD and $\D_1,\D_2 \in \Omega$.
Then, $\D_1 < \D_2$ (i.e., planning domain $\D_2$ is a refinement of domain $\D_1$) \emph{\textbf{iff}} there exists a planning domain $\D_2'$ such that: 
\begin{enumerate}
\item $\Executions(\D_2,s) = \Executions(\D_2',s)$, for all $s \in S$ (that is, $\D_2'$ are equivalent planning domains); and
\item $\D_2$ is an oneof-refinement of $\D_1$.
\end{enumerate}
\end{theorem}

This states that the only meaningful difference between ordered domains in $\Omega$ comes, only, in the refined domain ($\D_2$) having \emph{less} (in terms of set inclusion) non-deterministic effects in some operators.

\subsection{Compilation to Dual FOND planning}

Let $\M = \tuple{ \tuple{\Omega,\leq}, s_I,\G}$ be a a multi-tier planning problem such that $\D \leq \D'$ if and only if $\D'$ is a oneof-refinement of $\D$.
Due to Theorem~\ref{theo:nd-refinment_universal}, restricting $\leq$ to a oneof-refinement relation does not affect generality.
From now on, for technical legibility and without loss of generality, we assume domains in $\Omega$ are STRIP-like with no conditional effects.\footnote{All results can be generalized to domains with conditional effects, but would result in a significantly more cumbersome presentation and notation without providing significant insights.}

In this section, we shall construct a single dual-FOND planning problem $\P_\M = \tuple{\D_\M,s_\M,G_\M}$ that will fully capture problem $\M$.
For compactness, we use $\Eff^\D_o$ to denote the effects of operator $o$ in planning domain $\D$.
We also abuse notation and treat non-deterministic effects as sets.

\smallskip
Let us start by explaining the general strategy being encoded into $\P_\M$.
Roughly speaking, the planning problem $\P_\M$ will model a dynamic system running as per multi-tier specification $\M$.
As such, at any time, the system is operating relative to some model $\D$ in $\Omega$ (initially, the most ambitious $\DT$), trying to achieve $\D$'s goal via an appropriate plan, and degrading to an adequate (lower) model when action outcomes' do not align with model $\D$.
To achieve this, the encoding will model an \emph{iterative two-phase} process in which an \emph{acting} phase is, sometimes if necessary, followed by an \emph{alignment \& degradation} phase.
A special variable $\step$ is used to distinguish both phases.
As expected, during an \emph{acting phase}, an operator representing some domain action is executed.
This step involves the execution of a non-deterministic action with fair semantics, and the optional subsequent execution of an unfair version of the action. 
In the latter case, the system will then evolve to an \emph{alignment phase}, in which the encoding verifies whether the outcomes seen correspond to the assumed current model $\D$; and if not, the behavior is ``degraded'' to an appropriate (lower-level) model that is able to explain  the observed outcome.

It turns out that one of the key challenges is to encode a proper and scalable alignment phase in a planning domain (i.e., in PDDL): \emph{how can we encode that a given effect could be explained by some model in $\Omega$ (but not by another model)?}
In some sense, doing so would amount to reducing meta-level reasoning to the object (PDDL) level.
We show that, via a clever encoding, that reduction is indeed possible.
The technical difficulty is depicted in the following example.

\begin{example}
Consider the case in which the robot is operating in the highest domain model $\D_3$ and decides to execute action \pddl{walk}. 
Upon execution, the robot senses variable \pddl{scratch} true---the robot is now damaged.
In the standard (intuitive) configuration, in which the robot starts non-damaged, the robot should \emph{degrade} its operational model to domain $\D_2$, as that is the highest model explaining the damage.
However, if the robot happens to start damaged already, then domain model $\D_3$ still explains the transition, and no degradation should occur.
Here, \pddl{walk}'s effects under $\D_3$ and $\D_2$ are indistinguishable.
\label{example:explicability}
\end{example}

This example shows that just observing a proposition (\pddl{scratch}) in a transition that does not appear in an effect (\pddl{walk}'s effect under $\D_3$) does \emph{not} directly imply the effect may not explain the transition.
Can we then characterize, succinctly, under which conditions a set of observed propositions $E$ is explained by some operator $o$ in a domain model $\D$?
It turns out we can. 

\newcommand{\explains}{\mathname{Explains}}

\begin{definition}[Effect Explicability]\label{def:effect_explicability}
Let $E$ be a set of literals (e.g., effects that have just seen to be true after an action execution).
The conditions for operator $o$ in domain $\D$ to explain $E$, denoted $\explains[o,\D,E]$, is defined as follows (recall $\Delta$ is the set symmetric difference operation):
    \[
        \explains[o,\D,E]  = \bigvee_{E' \in \Eff_o^{\D}} \bigwedge_{l \in E \Delta E'} l.  
    \]
\end{definition}

\noindent
That is, some effect $E'$ of $o$ in model $\D$ yields the same result as effect $E$, if all the literals that one effect has an the other does not \emph{were already true} (at the outset of $o$'s execution).
In our Example~\ref{example:explicability}, if we take $E$ to be the second effect of $\pddl{walk}$ in $\D_2$ (i.e., $E = \set{\pddl{(not (at ?o))},\pddl{(at ?d)},\pddl{(scratch)}}$) we have $\explains[\pddl{walk},\D_3,E] = \pddl{scratch}$, as the literal \pddl{scratch} is the only one in the effects' symmetric difference. 

Observe that if $E$ and $E'$ are inconsistent, the formula will contain the conjunction of a proposition and its negation, thus reducing to false.
In fact, the following result guarantees the intended meaning of the above definition.

\begin{lemma}
Let $s,s' \in 2^{V \cup \overline{V}}$ be two domain states.
Let $E \in 2^{V \cup \overline{V}}  \supseteq s' \setminus s$ be a set of literals including at least all new literals in $s'$ w.r.t. to $s$.
Then, state $s'$ is a possible successor when operator $o$ is executed in state $s'$ under model $\D$ if and only if $s \models \explains[o,\D,E]$.
\end{lemma}

\bigskip
We are now ready to provide the encoding of $\M$ into a dual-FOND planning  problem $\P_\M = \tuple{\D_\M,s_\M,\G_\M}$.

\paragraph{Domain variables.}
The \emph{set of propositional variables} $V^+$ of $\D_\M$ is obtained by extending the set of variables $V$ in $\M$'s domains with the following additional variables:
\begin{itemize}
     \item $\varepsilon_\D$, for each domain $\D \in \Omega$, that will be used to signal that model $\D$ is a/the highest model explaining the effect of the last executed action;

     \item $\ell_\D$, for each domain $\D \in \Omega$, that will be used to track the most ``ambitious'' compatible model so far; 
     
     \item $\step$, use to denote the system is in the \emph{acting phase} (otherwise, it is in the \emph{alignment phase});
     
     \item $u_o$, for each operator $o \in \DT$, that will be used to ensure the execution of a unfair action; and
     
     \item $\pend$, used to denote the goal achievement of the current model of execution.
\end{itemize}

\paragraph{Initial state \& goal condition.}
The \emph{initial state} of $\P_\M$ is:
\[
    s_\M = s_I \land [\ell_{\DT} \land \bigwedge_{\D \in \Omega^-} (\neg \varepsilon_\D \land \neg \ell_\D) \land \step \land \neg \pend].
\]

This encodes the initial state $s_I$ of the \MTP $\M$ and the fact that the system starts in the $\Omega$'s greatest, most ambitious, domain model $\DT$ (proposition $\ell_{\DT}$ and all other $\ell_x$'s are false) and in the action phase.  
In addition, all effect level signaling variables $\varepsilon_x$ are initialized to false (no action has been executed), as well as the goal variable $\pend$. 

Finally, the goal condition of $\P_\M$ is simply $G_\M = \pend$.
We will see below which actions make variable $\pend$ true.

\vspace{-0.3cm}
\paragraph{Domain operators.}
The planning domain $\D_\M$ will include \emph{two types} of operators, one for modeling the actual domain actions and one for implementing the alignment check (and potential degradation) process.
Let us start with the former.

So, for each (domain) operator $o$ in domain $\D \in \Omega$, we include a operator $\tuple{o_\D,\Pre,\Eff}$ in $\D_\M$, where:
\begin{itemize}
     \item $\Pre = \Pre^{\D}_o \land \ell_\D \land \step \land \bigwedge_{o \in \DT} \neg u_o$, that is, action $o_\D$ is executable when $o$ itself is executable in $\D$, and the system is currently operating under model $\D$ and is the fair-acting phase ($\step$ is true and all $u_x$ are false); and
     
     \item $\Eff = \Eff^\D_o \cup \set{u_o}$, that is, when operator $o_\D$ is executed, either one of original effects of $o$ in $\D$ ensues or a distinguished predicate $o_u$ is made true. 
\end{itemize}

When one of the effects of $o$ in domain $\D$ happens, it just resembles the dynamics of domain $\D$.
However, if the effect that ensues is $u_o$, the system evolves into a ``unfair-acting" phase ($\step \land u_o$), explained after the following example.

\begin{example}\label{example:fair_action}
The resulting walk action for the domain level $\D_2$ in the compilation would look as follows in \PDDL:
{
\lstset{basicstyle=\small\ttfamily} 
\lstset{basicstyle=\scriptsize\ttfamily} 
\begin{lstlisting}
(:action walk_d2
  :parameters (?o - cell ?d - cell)
  :precondition (and (at ?o) (adj ?o ?d) (not (broken))
        (d2) (act) (not (u_walk)) (not (u_run)))
  :effect (oneof
   (and (not (at ?o)) (at ?d))
   (and (not (at ?o)) (at ?d) (scratch))
   (u_walk) ))
\end{lstlisting}
}
\end{example}

Next, when the the system evolves to the unfair-acting phase (e.g., due to effect \pddl{u_walk} happening in the example above), the only executable action will be a second version of domain operator $o$, which in turn will include \emph{all effects} of $o$ \emph{across all domains} in $\Omega$, together with additional book-keeping variables $\varepsilon_x$ to support the next alignment, and potential degradation, reasoning phase.
More concretely, for each domain operator $o$ in $\DT$, $\D_\M$ includes a rather powerful operator $\tuple{\ounf,\Pre,\Eff}$, where:
\begin{itemize}
     \item $\Pre = \step \land u_o \land \Pre_o$, that is, the system is in the unfair-acting phase for operator $o$; and
     
     \item $\Eff$ is a set of nondeterministic effects, each being a collection (i.e., conjunction) of conditional effects built as follows. 
     For every effect $E$ of operator $o$ that is mentioned in a domain $\D$ but not in any lower one (i.e., $E \in \Eff_o^\D \setminus \bigcup_{\D': \D' < \D} \Eff_o^{\D'}$), $\Eff$ contains, as one of its non-deterministic effects, the following complex effect:
     \[
         \displaystyle
           \bigwedge_{\D': \D' \geq \D}
            (\ceffect{C_{\D'}^E}{E \land \neg \step \land  \neg u_o \land  \varepsilon_{\D'}),  
            }
     \]
     where $C_{\D'}^E \! = \! \explains[o,\D',E] \land \!\!\!\!  
        \displaystyle \!\!\!\!\! 
          \bigwedge_{\D'': \D'' > \D'} \!\!\!\!\!\! \!\!\!\neg \explains[o,\D'',E]$.
     
    Intuitively, the operator $o_\unfair$ contains each possible effects $E$ of $o$ (from any domain in $\M$) as a non-deterministic option. 
    In turn, the set of conditional effects for a particular effect $E$ will not only make the effect $E$ itself ensue, but will also set a ``marker" proposition  $\varepsilon_\D$ signaling the highest domains explaining the effect in question.    
    To realize that, condition $C_{\D'}^E$ above states that the original effect $E$ is explained (as per~\Cref{def:effect_explicability}) by (some effect of) operator $o$ at domain model $\D'$ but not by any other model higher than $\D'$.
    When that is the case, proposition $\varepsilon_{\D'}$ is set to true, recording the fact that $\D'$ is the \emph{highest} model explaining such effect. Observe that by the way all conditions are designed, they ought to be mutually exclusive, so only one $\varepsilon_x$ will be made true.
    In addition,  $\step$ is set to false so as to force the reasoner into the \textit{alignment} phase, to be explained shortly.
    (We note that the effects of level $\D$ itself are accounted when $\D' = \D$.) 
\end{itemize}

Importantly, while $o_\D$ operator will be treated \emph{fair}, action $o_\unfair$ will be treated as \emph{unfair}---this is where dual FOND semantics~\cite{GeffnerGeffner:ICAPS18-FONDSAT} come into play.
Also, as the following example shows, significant syntactic simplifications can be achieved in $o_\unfair$ by analyzing conditions in conditional effects and precondition of the action.

\begin{example}
Let us see complete Example~\ref{example:fair_action} by showing the unfair version of the \pddl{walk} action. 
After syntactic simplification w.r.t. conditions and the action precondition, we obtain the simpler, more readable, equivalent action:
\begin{lstlisting}
(:action walk_unfair
  :parameters (?o - cell ?d - cell)
  :precondition (and (act) (u_walk)
                     (at ?o) (adj ?o ?d) (not (broken)))
  :effect (and (not (act)) (not (u_walk))   
    (oneof
      (when (true)  
          (and (not (at ?o)) (at ?d) (e3)) )
      (and (when (not (scratch))  
              (and (not (at ?o)) (at ?d) (scratch) (e2)) )
           (when (scratch)
              (and (not (at ?o)) (at ?d) (e3)) ))
      (when (true) (and (scratch) (e1) )) )))
\end{lstlisting}
\label{example:conditional_effects}
\end{example}

As we discussed before, this unfair action contemplates the effects present in all the domain models.
The intended meaning of this is that whenever an action executes, it may fail and we may observe effects of any domain level.
However, we do not want the planner to rely on these possible failures, so we contemplate them as unfair actions.

\paragraph{Alignment \& degradation operators.}
When the unfair version of a domain operator has been executed, an effect could ensue that might not be explained by the current domain under which the reasoner is operating under (encoded via propositions $\ell_x$). If so, the system ought to gracefully degrade to a lower level model that is able to explain the last system evolution.
We encode this reasoning, and potential degradation, in the so-called \emph{alignment} phase ($\step$ is false). 

In the best case, the state observed after the execution of an action corresponds to one of the expected ones w.r.t. the current planning domain model the executor is operating under.
Technically, the reasoner continues operating under current model $\D$ (proposition $\ell_\D$ is true), provided domain $\D$ has been able to explain the evolution of the last executed action: proposition $\varepsilon_{\D'}$ has been set to true for some domain $\D'$ that is either $\D$ itself or a higher one in the hierarchy (recall effects in higher level domains are subsets of  ).
So, in such case, the planner (and executor) is able to execute special action $\tuple{\continue_{\D}, \Pre,\Eff}$ to keep planning/executing under the current model and goal:

\begin{itemize}
     \item $\Pre = (\neg \step \land \ell_\D \land \bigvee_{\D' \geq \D} \varepsilon_{\D'})$, that is, the action can be executed during the alignment phase when the current domain or one of its refinements accounts for the last effect outcome.
     
     \item $\Eff = (\step  \land \bigwedge_{\D \in \Omega^-} \neg \varepsilon_\D)$, that is, effect signals are all reset and the system goes back to the action phase.
\end{itemize}

If, on the other hand, the state observed does \emph{not} conform to the current operating model (i.e., proposition $\varepsilon_\D$ is false), then the system must \emph{degrade} to a lower tier where the environment model would fit the observation, and adjust the objective to the corresponding (often less ambitious) goal. 
Needless to say, we expect a ``smart'' reasoner/executor to degrade as little as possible, by retaining as many assumptions on the environment as possible and only dropping those that have been observed to be wrong.
This will allow the agent to aim for the highest, most valuable, goal so far.

Technically, when $\D,\D' \in \Omega$ such that $\D' < \D$, we include an operator $\tuple{\degrade_{\D\D'}, \Pre,\Eff}$ in $\P_\M$, where:

\begin{itemize}
     \item $\Pre = \displaystyle
      \neg \step  \land \ell_\D \land 
      \!\!\!\!\!\!\!\!\!
        \bigvee_{\set{\D^*: \D^* \geq \D', \neg (\D \geq \D^* \geq \D'), \D^* \not \geq \D}} \!\!\! 
          \varepsilon_{\D^*}$; and

     \item $\Eff = 
        \neg \ell_\D \land \ell_{\D'} \land 
          \bigwedge_{x \in \Omega} (\neg \varepsilon_x \land \step)$.
\end{itemize}

That is, the controller can degrade from current operating domain $\D$ to domain $\D'$ if the last effect seen was explained by lower domain $\D'$ or any other domain higher than $\D'$ that is unrelated to $\D$ (so as to handle \MTP{}s with a non-linear structure).
The effect results in the controller being degraded to level $\D'$ (proposition $\ell_{\D'}$ becomes true), all booking explicability effect prepositions $\varepsilon_x$ being reset, and the reasoner progressing to the acting phase.

Note that, effectively, the dynamics of level variables $\ell_x$ are \textit{outside the control of the reasoner}, as these depend only on which non-deterministic effects have occurred and how (i.e., how variables $\varepsilon_x$ have been set). 

\paragraph{Goal operators.}
The only part remaining is the overall goal of the multi-tier problem.
Intuitively this should be ``achieve the highest level goal'', which under a conservative degradation process, it reduces to ``achieve the goal of the current operating model.'' 
We therefore include goal actions 
$\tuple{\checkgoal_{\D}, (\G_\D \land \ell_\D),\mathname{end}}$, one per domain $\D \in \Omega$.

\medskip
This completes the encoding of a multi-tier planning problem $\M$ into a single non-deterministic planning domain $\P_\M$.
We now prove its correctness w.r.t. Definition~\ref{def:mtc_solution}.
First, any solution policy for the planning task amounts, as is, to a solution to the corresponding multi-tier planning problem.

\begin{theorem}\label{def:soundness}
If $\pi$ is a \strongcyclic solution for $\P_\M$, then controller $\C^\pi(\D)$ is an \MTC solution for $\M$, where:
\[
    \C^\pi_{\D}(s) = \pi(s \land \ell_\D \land \step \land \bigwedge_{o \in \DT} \neg u_o)\text{, for all $s \in S$}.  
\]

\begin{proofsk}
Consider $s_i \in S$ and $\D \in \Omega$ such that $s_i \in \InitStates(\D,\C^\pi)$, and an infinite and fair execution $\lambda \in \Executions_{\C^\pi}(\D,s_i)$. We show that goal $\G(\D)$ holds true somewhere along $\lambda$ as follows:
\begin{enumerate}
  \item We transform $\lambda$ into an execution $\hat{\lambda} \in \Executions_{\C^\pi}(\D_\M,s_i^+)$, with $s_i^+ =  s_i \cup \set{\step,\ell_\D}$, by adding propositions $\step$ and $\ell_\D$ to every state in $\lambda$ and replacing every domain operator $o$ with its $o_\D$ version.

  \item If $so_\D$ appears infinitively often in $\hat{\lambda}$, we replace every second appearance of the form $s o_{\D} s'$ by two-action steps $s o_\D(s \cup \set{u_{o_\D}}) \ounf (s' \cup \set{\step,\varepsilon_{\D'}})$ such that $\D' \geq \D$ is the highest domain in $\Omega$ that contains the effect of $o$ that supports the transition $s o_{\D} s'$---we know there is one because $\hat{\lambda}$ is a legal execution in $\D_\M$ from state $s_i^+$.
  By doing this changes in $\hat{\lambda}$ we are guarantee that the execution is fair, while still preserving the fact that every domain action in it has the effects as per domain $\D$. So, execution $\hat{\lambda}$ mirrors the original $\lambda$ for domain $\D$ but over the extended language of $\D_\M$.
  
  \item Because $s_i \in \InitStates(\D,\C^\pi)$, there exists a finite execution $\lambda_i \in \Executions_{\pi}(\D_\M,s_I)$ (i.e., execution in $\P_\M$ via policy $\pi$) that ends in state $s_i \cup \set{\ell_\D, \step}$.
  This means that $\lambda_i\hat{\lambda} \in \Executions_{\C^\pi}(\D_\M,s_I)$, and since $\lambda_i\hat{\lambda}$ is fair (w.r.t. the fair actions) and $\hat{\lambda}$ has $\lambda_\D$ always true, it follows that $\hat{\lambda}$ has to reach the $\P_\M$'s goal by executing operator $\checkgoal_{\D}$. Then, its precondition $\G_\D$ holds true at some point in $\hat{\lambda}$ and therefore in $\lambda$ too.
\end{enumerate}
\end{proofsk}
\end{theorem}

That is, the \MTC controller $\C$ under domain $\D$ and in state $s$, what the strong-cyclic solution for $\P_\M$ prescribes when $\ell_\D$ is true and the reasoning cycle is in the acting phase.

In addition, any possible \MTC solution will be represented by some \strongcyclic policy of $\P_\M$ (i.e., completeness).

\begin{theorem}\label{theo:complete}
If $\C$ is an \MTC solution for $\M$, then there exists a \strongcyclic solution $\pi$ for $\P_\M$ such that $\C^\pi(\D) = \C(\D)$, for every domain $\D$ in $\M$ (where $\C^\pi(\D)$ is as in \Cref{def:soundness}).

\begin{proofsk}
Policy $\pi$ follows the domain actions prescribed by $\C_\M$ exactly, augmented with the booking auxiliary actions as needed.  A similar argument, based on execution traces, as in Theorem~\ref{def:soundness} can be built.
\end{proofsk}  
\end{theorem}

We close by noting that the size of $\P_\M$ is increased by a linear number of bookkeeping propositional variables, and a quadratic number (w.r.t. the number of domains in $\Omega$) of extra actions.
So, while the multi-tier planning framework appears to be more involved than the standard (non-deterministic) planning, it can be suitably reduced to the latter, with an encoding that is, arguably, fairly natural and comparable in size.
Importantly, though, the solution proposed relies on the fact that we can specify \textit{both} fair and unfair actions in the same planning model. This is a feature that will prove a challenge when actually realizing the technique in current planning technology, as we shall see next.



\section{Validation and Discussion}\label{sec:validation}

In this section, we demonstrate that {\MTP}s can indeed be solved \textit{today} with existing planning technology, but argue that additional effort in Dual FOND is necessary.
The first obstacle is the availability of FOND planning technology supporting both fair and unfair assumptions.
To the best of our knowledge, the only off-the-shelf planner to do so is \citeby{GeffnerGeffner:ICAPS18-FONDSAT}'s FOND-SAT system.
By leveraging on SAT solvers, their system yields an elegant declarative technique for FOND planning that features the possibility of combining fair and unfair actions. 
So, we report on using FOND-SAT over the encoding for our non-running example.
Notwithstanding, the experiments reported are intended to demonstrate the existence of systems to solve \MTP{}s and to provide a baseline for future work, rather than for providing a performance evaluation.

%
\begin{listing}
\lstset{numbers=left, numberstyle=\tiny, stepnumber=5,%
numberfirstline=false, numbersep=5pt}
  \begin{lstlisting}[morekeywords={[2]plan},morekeywords={[6]case,else,if},stepnumber=1]
  (:plan [
    (walk_d3 c2 c1)
    (if ((not (u_walk))) [
      (walk_d3 c1 c0)
      (if ((not (u_walk))) [
        (check_goal_d3)
      ])
      ... 
    ]
    (else) [
      (walk_unfair)
      (case (eff_e3_walk) [
        (walk_e3_explained_by_d3)
        (continue_d3)
        ...
      ]
      (case (eff_e2_walk) [
        (walk_e2_explained_by_d2)
        (degrade_d3_d2)
        ...
      ]
      (case (eff_e1_walk) [
        (walk_e1_explained_by_d1)
        (degrade_d3_d1)
        ...
      ]
    ]  ])
  \end{lstlisting}
  \vspace*{-0.5cm}
  \caption{A fragment of the policy found by FOND-SAT.}
  \label{fig:FONDSAT-policy}
\end{listing}



\smallskip
\Cref{fig:FONDSAT-policy} shows a fragment, in a readable plan-like format, of the outcome when FOND-SAT is ran on our encoding for the non-running example. The full controller is shown in Appendix~\ref{apx:non-running}.\footnote{The code used to perform the compilation from the multi-tier specification to Dual-FOND can be found here: \url{https://github.com/ssardina-planning/pypddl-translator}}.
First of all, the plan cautiously avoids the run action altogether, as it may get the robot broken and precludes the achievement of all tier goals.

After performing the walk \textit{fair}-version action in (line 2) corresponding to the highest model $\D_3$, the plan checks its effects (line 3).
If proposition $\pddl{u_walk}$ remains false (lines 4-9), the effect in model $\D_3$ has occurred---the robot has done a successful move. If another walk (line 4) succeeds as well (lines 5-7), the robot achieves the top level $\D_3$' goal (line 6).
Note that, in such a run, no alignment action is included: the walk unfair version has never been performed and hence only effects of $\D_3$ has ensued.

If, instead, the first walking action (line 2) yields the special effect $\pddl{u_walk}$, the plan jumps to line 11.
There, the only action available is the \textit{unfair} version of walking (line 11), which has \textit{all} the effects, as non-deterministic options, of the walking action across all domains $\D_3$, $\D_2$, and $\D_1$.
As FOND-SAT does not handle conditional effects, we simulate each conditional effect for the effect chosen by a set of \pddl{walk_eE_explained_by_dx} whose precondition is $C^E_{\D_x}$, together with the original precondition of the operator (\pddl{walk} in this case).
Finally, if the effect chosen could be explained by the current operating domain (line 13, explained by domain $\D_3$), the system executes a continue operation at the current level, enabling the next domain action.
On the other hand, when the effect is explained by a lower domain than the one operating under (lines 18 and 23), degradation to the corresponding domain is carried out (line 19 and 24).

It is easy to see how this plan also represents a multi-tier controller which only outputs the  fair version of the operators, and all the other auxiliar actions and propositions are the controller internal memory. 

Now, what would happen if the robot starts scratched as discussed in Example~\ref{example:explicability}?
It turns out the problem becomes \emph{unsolvable}. The reason is that any observed scratch after movement does \emph{not} need to be explained by a different model than $\D_3$ (e.g., the walk effect of $\D_2$ that scratches the robot), as the scratch is explained already by being true originally. Thus, when walking always advances the agent, it would never degrade its behavior, remain operating in $\D_3$ without ever achieving $\D_3$'s goal.
If, however, we drop the non-scratched requirement from $\D_3$, the problem would be solvable again, though with a slightly different policy.  The robot would just try to achieve the top goal, degrading only to $\D_1$ if it does not move after a walk action.
Since, as discussed, $\D_2$'s scratch effect would be explained by $\D_3$ itself, line 18 would become \pddl{walk_e2_explained_by_l3} and line 19 would become \pddl{continue_l3}. The policy for this example is shown in Appendix~\ref{apx:non-running}.

\medskip
While the above demonstrates the possibility to solve \MTP{}s using (the only) existing planning technology, running our, arguably simple, example takes around 600 seconds to produce the 29 states first controller shown in the Appendix~\ref{apx:non-running} in an i7-4510 CPU with 8GB of RAM, when using the off-the-shelf version of the planner. 
This clearly indicates the need for more and better dual-FOND implementations or the development of specialized optimizations for \MTP{}s.
For example, as we are only allowing degradation and not upgrades, one can modify the SAT encoding to specify a number of controllers to be used per domain level, without allowing transitions from lower to upper levels. 
In preliminary tests we did we experienced a speed-up of approx 30\%. 
Another optimization involves an estimation of the number of controllers required to solve the \MTP, for example by running the top level domain which will provide a lower bound.





\section{Related work}

The work presented in this paper is mostly related to existing work that aim to better handle execution failures and exceptions as well work that aim to provide richer goal specifications.

Fault Tolerant Planning (FTP) explores how to build plans that can cope with operations' failures. 
Notably, the work of~\citeby{Domshlak:ICAPS13} considers the FTP problem via reductions to planning. 
There, $k$-admissible plans are defined as those that guarantee to achieve the goal even if an operation happens to fail up to $k$ times. 
Like ours, the approach reduces to (classical) planning via an intelligent compilation inspired in that of~\cite{BonetGeffner:IJCAI11}.
In other words, the problem is reduced to bounded liveness. 
We do not impose such a restriction and our adaptive framework is actually orthogonal to theirs. 
In fact, it would be possible to accommodate $k$-admissible plans within our hierarchy, so that degradation is trigger only after some number of observed failures.

In Robust Planning (RP), the usual approach is to search for the best plan assuming the worst possible environment, e.g.,~\cite{BuffetAberdeen:IJCAI05}. 
Typically, RP looks for subclasses of MDPs for which policies are guaranteed to eventually reach a goal state. Policies are normally computed as solutions to Stochastic Shortest Path problems.
Our approach, instead, is based on a \emph{qualitative} formalization of action and change, with no specification of probabilities, which can sometimes be not trivial, unfeasible, or simply uneconomical to obtain.
More importantly, having a single problem description forces the engineer to represent a goal that may never be achieved in the actual environment. Indeed, the engineer may find it difficult, or impossible, to model degraded goals depending on the actual environment's dynamics; it would require the goals to somehow predicate on the probabilities associated to the assumed degraded behavior. 
We in turn provide an accessible formalism for modelling different levels of assumptions and goals.
Also, our technique is not rooted in dynamic programming or optimization algorithms, but in planning ones.

In terms of representation formalisms, there has been efforts to provide more powerful goal specifications, such as 
\citeby{LagoPistoreTraverso:AAAI02}'s \textsc{EaGle} language, ~\citeby{DeGiacomo.etal:AIJ16}'s Agent Planning Programs, and \citeby{Shivashankar.etal:AAMAS12}'s Hierarchical Goal Networks (HGN), but always assuming operation on a single model of the environment aiming for a single goal. 


Our work also keeps some relationship with deliberative control architectures such as T-REX~\cite{DBLP:conf/icra/McGannPRTHM08}, and other similar planning-execution-monitoring architectures like Propice-Plan~\cite{DBLP:conf/ecp/DespouysI99} or CPEF~\cite{DBLP:journals/aim/Myers99}. 
Our approach differs in that those systems usually \emph{replan} when goals (or perceived state) change, whereas our proposal aims to generate a policy that takes into account all the scenarios \emph{prior} to the system's execution (as per model). 
As a result, we can provide guarantees of goal achievability and program termination (although it s more demanding computationally). 
Also, our framework puts more emphasis on formal semantics (based on PDDL and transition systems), modeling, and plan synthesis, whereas the mentioned agent architectures primarly focus on implemented platforms and execution.

As stated, our work is inspired by that of~\citeby{DIppolito.etal:ICSE14} in the area of Software Engineering.
At a general level, our account is rooted in Knowledge Representation, and specifically, automated planning, which allows us to leverage on advanced representation formalisms (e.g., \PDDL) as well as computational techniques for such representations.
Still, as ours, their approach provides support for a tiered architecture with multiple behavioural models and goals.
Unlike ours, though, their account is limited to a lineal hierarchy of models, so independent assumptions, as in our example, cannot be represented.
More importantly, \citeauthor{DIppolito.etal:ICSE14} require solution (sub-)controllers of lower tiers to \emph{simulate} those in upper tiers, and thus it would not handle our simple no-running example.
Finally, being rooted in knowledge representation, we area able to exploit planning technology.

\section{Conclusions}\label{sec:discussion}

The overarching motivation of this work is to addresses~\citeby{Ghallab.etal:AIJ14}'s call for an \emph{actors' perspective} on planning, by proposing a pure planning-based framework that ``integrates better planning and acting.'' 
%
In our framework, the knowledge engineer has the opportunity to consider multiple levels of assumptions and goals. 
The problem amounts to synthesize a meta-controller that, while running, is able to gracefully degrade its ``level of service'' when the assumptions on the environment are not met. 
We developed a compilation technique to construct such adaptive meta-controllers via dual-FOND planning. We note that plain FOND planning, where every action is assumed fair, would not work, as the agent may decide to keep trying an action to obtain one of the ``failing" effects to achieve an ``easier" lower-level goal (this artifact was already noted by~\citeby{CamachoMcIlraith:KNOWPROS16}).


There are several limitations of our framework in its current form. 
The approach is based on lattice structures that do not guarantee full ordering of environment models.
As a consequence, the executor may have multiple optimal options to degrade execution from a given tier, and it has to ``guess'' one.
But, upon a later inconsistency, the executor may find out it should have degraded differently.
Since, the proposal does not provide support to move ``sideways'' in the same tier, the executor is forced to degrade downwards again. 
Also, we have not provided a mechanism for \emph{enhancement}, that is, for ``upgrading'' to more refined models, for example, when certain transient failure has been fixed.


\section*{Acknowledgements}
Alberto Pozanco carried out this work during his visit to RMIT University supported by FEDER/Ministerio
de Ciencia, Innovaci\'on y Universidades – Agencia Estatal ´
de Investigaci\'on TIN2017-88476-C2-2-R and RTC-2016-5407-4.

\newpage

\bibliographystyle{aaai}

\newpage
\onecolumn
\setcounter{secnumdepth}{1}
\renewcommand\thesection{\Alph{section}}

\section{Non-Running Example}\label{apx:non-running}

The following is graphical representation of the policy obtained for the non-running scenario:
\begin{center}
    \includegraphics[width=.83\textwidth]{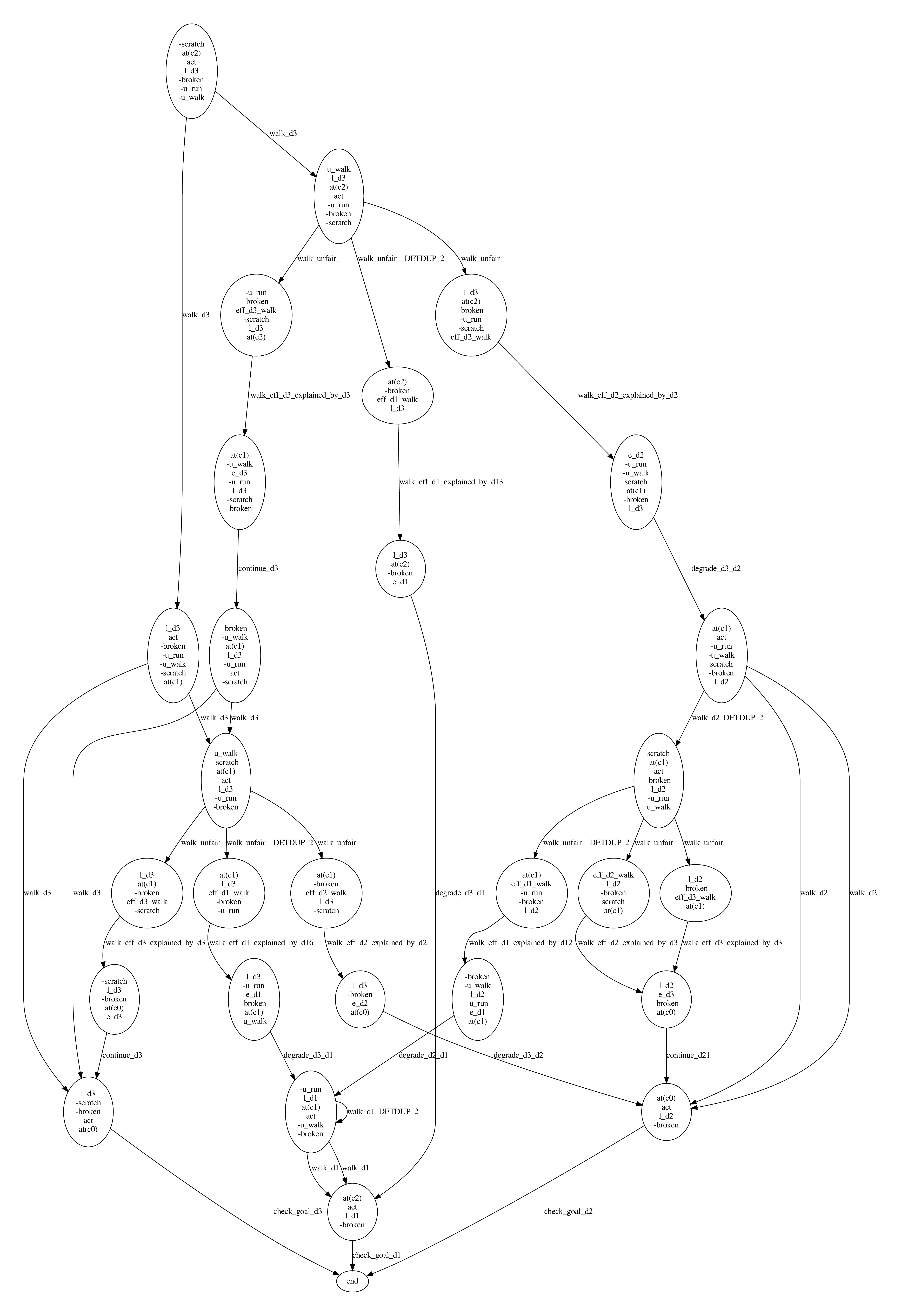}
\end{center}

In turn, the controller for the variant of the scenario where (1) the robot is initially scratched; and (2) the non-scratched requirement from $\D_3$'s goal is dropped is as follows:
\begin{center}
    \includegraphics[height=.95\textheight]{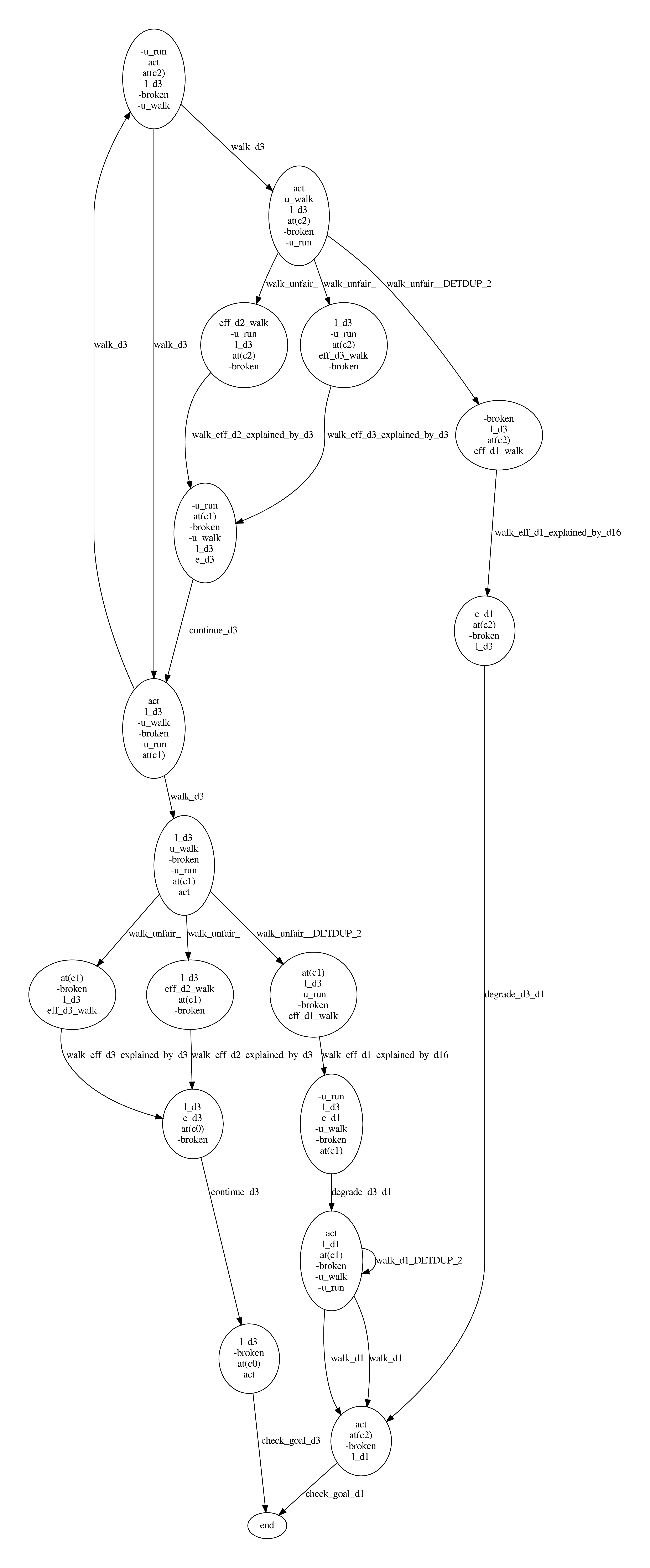}
\end{center}




\end{document}